\documentclass[letterpaper]{article} 
\usepackage{aaai2027}  
\usepackage[hyphens]{url}  
\usepackage{graphicx} 
\usepackage{natbib}  
\usepackage{caption} 
\usepackage{amsfonts} 
\usepackage{algorithm}
\usepackage{algorithmicx}
\usepackage{algpseudocode}
\usepackage{multirow}

\usepackage{newfloat}
\usepackage{listings}
\DeclareCaptionStyle{ruled}{labelfont=normalfont,labelsep=colon,strut=off} 
\floatstyle{ruled}
\newfloat{listing}{tb}{lst}{}
\floatname{listing}{Listing}

\usepackage{booktabs}
\usepackage{tikz}
\usetikzlibrary{arrows.meta, positioning}
\usepackage{amsmath}
\title{AAAI Press Anonymous Submission\\Instructions for Authors Using \LaTeX{}}
\title{Root Cause Analysis with Latent Confounders using Partial Ancestral Graphs}
\author {
    Henrique O. Caetano\textsuperscript{\rm 1}\corresponding,
    Rafael Arone\textsuperscript{\rm 1},
    Carlos Dias Maciel\textsuperscript{\rm 2}
}
\affiliations {
    \textsuperscript{\rm 1}Signal Processing Laboratory, São Carlos School of Engineering, University of São Paulo, São Paulo, Brazil\\
    \textsuperscript{\rm 2}Guaratinguetá School of Engineering, São Paulo State University, Guaratinguetá, Brazil\\
    \{henriquecaetano1, rafael.arone\}@usp.br, carlos.maciel@unesp.br 
}

\begin{document}
\nocopyright 
\maketitle

\begin{abstract}

Finding the source of failures, known as Root Cause Analysis (RCA), is essential for identifying the root causes of anomalies and maintaining the reliability of complex systems. While causal theory has advanced data-driven RCA, existing frameworks assume causal sufficiency, failing to account for the unobserved latent variables prevalent in real-world environments. To address this gap, we propose PAG-RCA. This framework models system failures as parametric interventions over Partial Ancestral Graphs (PAGs) to perform RCA in the presence of latent variables. We use standard causal identification algorithms to find the source of failures by quantifying causal effects over the PAG. When an effect is identifiable, candidate root causes are ranked based on their exact intervention effects. When effects are structurally unidentifiable, our framework (for the first time in the RCA literature) integrates partial identification to evaluate and score candidates using analytical causal bounds. By integrating latent variables and partial identification at once our framework ensures robust RCA even under data scarcity and latent-variable scenarios where traditional methods degrade. Evaluations on synthetic data, microservice anomaly benchmarks and power-grid cascading failures demonstrate that PAG-RCA consistently outperforms state-of-the-art data-driven baselines. By improving data-driven RCA performance under data scarcity, this methodology advances reliable automated diagnostics in partially observable complex networks.

\end{abstract}



\section{Introduction}
\label{sec:introduction}

\begin{figure*}[h!]
\centering
\resizebox{\textwidth}{!}{%
\begin{tikzpicture}[
    >={Stealth[length=2.5mm]},
    var/.style={circle, draw, inner sep=1pt, minimum size=0.65cm},
    latent/.style={circle, draw, dashed, fill=gray!10, inner sep=1pt, minimum size=0.65cm}
]

\begin{scope}[xshift=0cm]
    \node[var] (F) at (0,0) {$F$};
    \node[var] (X1) at (1.2,0) {$X_1$};
    \node[var] (X2) at (2.4,1) {$X_2$};
    \node[var] (X3) at (2.4,-1) {$X_3$};
    \node[latent] (L) at (3.6,0) {$L$};
    \node[var] (X4) at (3.6,1) {$X_4$};

    \draw[->] (F) -- (X1);
    \draw[->] (X1) -- (X2);
    \draw[->] (X1) -- (X3);
    \draw[->, dashed] (L) -- (X2);
    \draw[->, dashed] (L) -- (X3);
    \draw[->] (X2) -- (X4);
    
    \node[font=\bfseries] at (1.8, -2) {(a) True DAG ($\mathcal{D}$)};
\end{scope}

\begin{scope}[xshift=5.8cm]
    \node[var] (F) at (0,0) {$F$};
    \node[var] (X1) at (1.2,0) {$X_1$};
    \node[var] (X2) at (2.4,1) {$X_2$};
    \node[var] (X3) at (2.4,-1) {$X_3$};
    \node[var] (X4) at (3.6,1) {$X_4$};

    \draw[->] (F) -- (X1);
    \draw[->] (X1) -- (X2);
    \draw[->] (X1) -- (X3);
    \draw[{Stealth[length=2.5mm]}-{Stealth[length=2.5mm]}] (X2) -- (X3);
    \draw[->] (X2) -- (X4);
    
    \node[font=\bfseries] at (1.8, -2) {(b) MAG ($\mathcal{M}$)};
\end{scope}

\begin{scope}[xshift=11.6cm]
    \node[var] (F) at (0,0) {$F$};
    \node[var] (X1) at (1.2,0) {$X_1$};
    \node[var] (X2) at (2.4,1) {$X_2$};
    \node[var] (X3) at (2.4,-1) {$X_3$};
    \node[var] (X4) at (3.6,1) {$X_4$};

    \draw[->] (F) -- (X1);
    \draw[{Circle[open, length=1.5mm]}-{Stealth[length=2.5mm]}] (X1) -- (X2);
    \draw[{Circle[open, length=1.5mm]}-{Stealth[length=2.5mm]}] (X1) -- (X3);
    \draw[{Stealth[length=2.5mm]}-{Stealth[length=2.5mm]}] (X2) -- (X3);
    \draw[{Circle[open, length=1.5mm]}-{Circle[open, length=1.5mm]}] (X2) -- (X4); 
    
    \node[font=\bfseries] at (1.8, -2) {(c) PAG ($\mathcal{P}$)};
\end{scope}

\end{tikzpicture}%
}
\caption{Illustrative example of causal graphs. (a) A true DAG with a latent confounder $L$. (b) The corresponding MAG marginalizing $L$. (c) The PAG representing the Markov equivalence class learned from observational data.}
\label{fig:running_example}
\end{figure*}
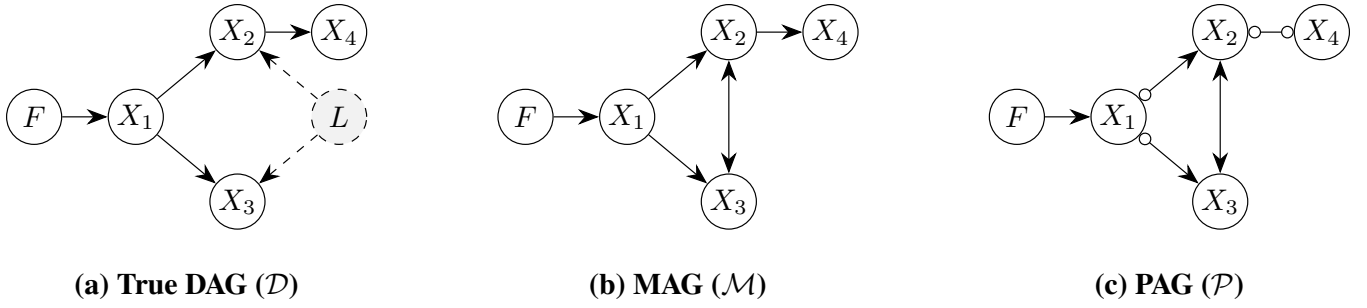

Root Cause Analysis (RCA) is the process of identifying the initial source of an anomaly or failure to enable system recovery. Automated RCA is required to maintain the reliability of networked systems across diverse domains, including power system cascading failures \cite{caetano2026spatiotemporal}, microservice architectures \cite{rcd_paper_2022,baro_2024_paper}, industrial transportation systems \cite{chen2025causal}, membrane fuel cells \cite{zuo2025degradation}, and medicine \cite{rani2026associated}. Modern environments consist of interdependent components where a single fault can propagate and disrupt the entire network. Manual diagnosis in these settings is time-consuming and often results in system downtime, service-level agreement violations, and financial loss. Consequently, developing automated methodologies to locate root causes is a primary objective for system management and operations \cite{baro_2024_paper,orchard2026root,rcg_paper_2025,cyclic_rca_paper_2026,aerca_paper_2025,easy_rca_paper}.

To address this challenge, researchers have proposed diverse techniques, including Bayesian change point detection \cite{baro_2024_paper}, machine learning models to define dependency relationships \cite{coral_paper_2023,diag_fusion_paper_2023,zheng2024mulan,run_algorithm_2024,kgroot_paper_2024,holistic_rca_2024}, extreme value statistical theory for outlier detection \cite{orchard2026root,shan2019diagnosis,rehak2023counterfactual,li2022causal}, and large language models \cite{tamo_paper_2025}. However, recent literature increasingly utilizes causal theory to formally compute cause-and-effect relationships and model system failures as parametric interventions \cite{easy_rca_paper,rcd_paper_2022,rcg_paper_2025,run_algorithm_2024,aerca_paper_2025}. These causal RCA frameworks generally operate through a two-phase process. First, a causal graph is learned from system data or known beforehand, where nodes represent metrics and directed edges encode the direction and magnitude of causal dependencies. In the second phase, candidate variables are ranked based on scoring functions evaluated over the constructed graph. Prominent ranking scores include hypothesis tests of outliers \cite{budhathoki2022causal}, the quantification of direct causal effects \cite{easy_rca_paper}, and conditional mutual information \cite{rcd_paper_2022,rcg_paper_2025}.

Some causal approaches for RCA rely on the assumption that the underlying causal graph is fully known a priori \cite{orchard2026root,budhathoki2022causal}, which is rarely feasible in complex, real-world systems. Instead, a more prominent direction is to employ a fully data-driven approach, where the causal graph is learned from observational data and subsequently used to rank root causes. For instance, the authors of \cite{easy_rca_paper} identify root causes by quantifying direct causal effects between normal and anomalous regimes given an acyclic summary causal graph. Other frameworks, such as \cite{rcd_paper_2022}, introduce an auxiliary F-node to model the failure as an intervention, effectively separating normal and abnormal operational regimes. A fundamental limitation of these data-driven methods is the assumption that causal discovery algorithms output a fully specified Directed Acyclic Graph (DAG). In practice, data is often insufficient to evaluate all conditional independencies, yielding a Markov Equivalence Class (MEC) that represents multiple possible valid graphs. Recent work has begun to address this structural uncertainty; notably, \cite{rcg_paper_2025} extends the F-node framework to account for MECs by evaluating Completed Partially Directed Acyclic Graphs (CPDAGs), highlighting the necessity of incorporating partial structures when full DAGs are unobtainable.

Despite these advancements, existing frameworks exhibit two critical gaps. First, the RCA literature relies on the assumption of causal sufficiency, assuming no latent confounders exist. In complex systems, latent variables are prevalent due to unmonitored external factors (e.g., varying workloads) or incomplete sensor coverage \cite{caetano2026spatiotemporal}. While authors in \cite{rcg_paper_2025} represents structural uncertainty via CPDAGs, properly accounting for MECs under hidden confounding requires modeling the system through Partial Ancestral Graphs (PAGs). Second, although current methods utilize causal theory for structure learning, they rank candidate nodes using correlation-based or associational scores, such as outlier hypothesis tests \cite{budhathoki2022causal} or conditional mutual information \cite{rcd_paper_2022,rcg_paper_2025}. The exact quantification of causal effects has been utilized \cite{easy_rca_paper}, but under the assumptions of a single known DAG and causal sufficiency. 

Combining the presence of latent confounders with the exact quantification of causal effects introduces significant methodological challenges. It necessitates the application of PAG-calculus \cite{jaber2022causal}, which remains unexplored in the RCA literature. Furthermore, in the presence of latent variables, specific causal effects may not be point-identifiable over the PAG. In such scenarios, causal bounds \cite{bellot2024towards} are required to establish a range of possible values for the interventional distribution. If the causal effect of the true root cause is unidentifiable, standard algorithms fail to locate the failure; bounding ensures that every effect is at least partially identifiable, allowing to rank all variables.

\textbf{Main Contribution.} We propose PAG-RCA, a novel framework designed to address the challenge of root cause isolation under hidden confounding. To the best of our knowledge, this is the first methodology to conduct data-driven RCA while explicitly accounting for the presence of unobserved latent variables. Our specific contributions are:

\begin{itemize}
    \item We model system failures as parametric interventions and introduce a framework to learn PAGs from normal and anomalous observational data, capturing structural uncertainty and unobserved latent variables.
    \item Previous data-driven frameworks often rely on correlational scores because causal effects are frequently unidentifiable due to latent confounding. We address this by establishing a fully interventional ranking mechanism. We compute interventional distributions via the CIDP algorithm when possible; when point estimation is infeasible over the PAG, we integrate partial identification and causal bounding. This ensures that every candidate can be scored and ranked, even under unidentifiability.
    \item We validate PAG-RCA through experiments on synthetic data, microservices, and simulated cascading failures in power grids. Our method demonstrates improved Top-$k$ accuracy compared to state-of-the-art baselines, especially under conditions of data scarcity, structural uncertainty, and latent confounding.
\end{itemize}

\section{Background}
\label{sec:background}

This section defines the core concepts utilized in our methodology. For formal treatments and extended details, please refer to the Extended version.

\textbf{Causal graphs and d-separation.} A causal graph encapsulates relationships using a Directed Acyclic Graph (DAG), where a directed edge $X \rightarrow Y$ indicates that $X$ causes $Y$. In Figure \ref{fig:running_example}a, the system operates under a DAG $\mathcal{D}$ where the intervention node $F$ causes $X_1$, and $X_1$ causes $X_2$ and $X_3$. Conditional independence in a causal graph is determined by d-separation. A set of vertices $Z$ d-separates $X$ and $Y$, denoted $(X \perp\!\!\!\perp Y \mid Z)_{\mathcal{D}}$, if $Z$ blocks all paths between them (i.e., every non-collider on the path is not in $Z$, and every collider is an ancestor of a node in $Z$).

\textbf{Markov equivalence and CPDAGs.} Different DAGs yielding the identical set of d-separation statements form a Markov equivalence class (MEC). This class is represented by a Completed Partially Directed Acyclic Graph (CPDAG). A CPDAG maintains the skeleton of the class, placing a directed edge $X \rightarrow Y$ if the direction is invariant across all DAGs in the class, and an undirected edge otherwise.

\textbf{Latent confounders and MAGs.} Unobserved variables affecting multiple observed variables are latent confounders (e.g., $L$ in Figure \ref{fig:running_example}a). Because DAGs and CPDAGs assume causal sufficiency, marginalizing these confounders produces a Maximal Ancestral Graph (MAG). A MAG represents confounding via bidirected edges, such as $X_2 \leftrightarrow X_3$ in Figure \ref{fig:running_example}b.

\textbf{Partial Ancestral Graphs (PAGs).} A PAG represents a MEC of MAGs using three edge endpoints to denote uncertainties: arrowheads ($>$), tails ($-$), and circles ($\circ$). As shown in Figure \ref{fig:running_example}c, a PAG contains four primary edge types. A directed edge ($X \rightarrow Y$) indicates $X$ is an ancestor of $Y$ in all MAGs within the equivalence class. A bidirected edge ($X \leftrightarrow Y$) denotes unmeasured confounding where neither node is an ancestor of the other. A partially directed edge ($X \circ\rightarrow Y$) specifies $Y$ is not an ancestor of $X$, but it remains undetermined if $X$ causes $Y$ or if they share a confounder. Finally, an undirected edge with circle marks ($X \circ-\circ Y$) indicates the ancestral relationship is entirely undetermined.

\textbf{Ancestrality and possible parents.} Ancestral bounds in a PAG are inferred directly from edge marks. Node $Y$ is a descendant of $X$ if there exists a directed path from $X$ to $Y$ consisting exclusively of $\rightarrow$ edges. Node $X$ is a possible parent of $Y$, denoted $PossPa(Y)$, if there is a path from $X$ to $Y$ with no arrowheads pointing towards $X$. In partial structures, possible parents are the variables acting as parents in at least one valid graph in the equivalence class (e.g., incident edges like $X \rightarrow Y$ or $X \circ\rightarrow Y$).

\textbf{Interventions and identifiability.} An intervention changes the generative mechanism of a variable. This is captured by the $do$-operator, where $do(X=x)$ forces a variable $X$ to take value $x$ \cite{bareinboim2022pearl}. Identifiability determines whether a post-intervention distribution $P(Y \mid do(X=x))$ can be computed from an observational distribution (i.e. $P()$ expression without the do() operator) and a causal graph. For DAGs, the ID algorithm \cite{id_algorithm} provides a sound and complete method to reduce an interventional distribution into a do-free expression. For PAGs, which accommodate unobserved confounders, the IDP (for marginal effects) and CIDP (for conditional effects)  are sound and complete algorithms for deriving do-free expressions \cite{jaber2022causal}.

\textbf{Causal bounds.} When a causal effect is not point-identifiable due to structural uncertainties in a PAG, we can account for partial identifiability. Observational data and the invariant properties of the equivalence class restrict the range of possible values. By calculating \textit{causal bounds}, we define an interval $[L, U]$ guaranteed to contain the true effect across all compatible structural models. For PAGs, the Partial CIDP algorithm \cite{bellot2024towards} derive these bounds from observational distributions and the PAG structure.

\section{Problem Statement}

Let $V = \{X_1, X_2, \dots, X_n\}$ be the set of observed continuous metrics in a complex system. We assume access to an observational dataset $D$, collected during the system's normal operation, and a dataset $D^*$, collected during an anomalous event. The primary goal of RCA is to identify the specific variable $R \in V$ (the root cause) that triggered the abnormal condition. Following previous literature \cite{easy_rca_paper,rcd_paper_2022,rcg_paper_2025,aerca_paper_2025}, we aim to learn cause-and-effect relationships among system metrics to isolate $R$.

\textbf{Failure as interventions.} We model a system failure as a soft intervention on the root cause metric. To formalize this, we introduce an auxiliary binary variable, the F-node ($F$), which researchers have used to represent the effect of interventions on a system \cite{yang2018characterizing, mooij2020joint}, and more recently in root cause analysis frameworks \cite{rcd_paper_2022, rcg_paper_2025}. We concatenate the datasets and set $F=0$ for samples from $D$ and $F=1$ for samples from $D^*$. This allows us to sample from a joint distribution $P^*$ where $P^*(V \mid F=0) = P_N(V)$ and $P^*(V \mid F=1) = P_A(V)$. Under this formalism, a variable $R \in V$ is defined as the root cause if its generative mechanism changes after the Failure, yielding the distributional variance $P_N(R \mid Pa(R)) \neq P_A(R \mid Pa(R))$. The conditional dependence $R \not\perp\!\!\!\perp F \mid Pa(R)$ translates into the augmented causal graph. Since the Failure affects the system but does not cause it, $F$ has no incoming edges.

\textbf{Data-driven causal discovery and equivalence classes.} While some RCA literature assumes a fully known DAG is available beforehand \cite{orchard2026root, cyclic_rca_paper_2026}, such perfect structural knowledge is rarely accessible in real applications. Consequently, data-driven frameworks learn the causal graph directly from data to conduct RCA \cite{easy_rca_paper, rcd_paper_2022, rcg_paper_2025, aerca_paper_2025}. Many existing frameworks assume that causal discovery algorithms output a fully specified DAG, failing to account for MECs. In reality, algorithms relying on conditional independence tests can only identify a causal structure up to its MEC \cite{jaber2022causal}. As a result, their output is a partial causal structure that encodes structural uncertainties (such as CPDAGs and MAGs), rather than the exact underlying DAG.

\textbf{RCA with latent variables.} Recent work has begun to address data driven causal discovery with MECs in RCA. The RCG framework \cite{rcg_paper_2025} was the first to explicitly account for MECs by evaluating CPDAGs, which represent the MECs of DAGs. However, CPDAGs operate under the strict assumption of causal sufficiency, meaning they assume that no latent variables exist in the data. In complex systems, latent variables are prevalent due to unmonitored external factors (e.g., weather, varying workloads) or incomplete sensor coverage. Our goal is to develop a fully data-driven framework capable of performing RCA in the presence of latent variables. To achieve this, we shift the underlying graphical representation from DAGs and CPDAGs to MAGs (Figure \ref{fig:running_example}b) and their corresponding MECs, PAGs (Figure \ref{fig:running_example}c). By leveraging PAGs, our goal is to account for latent variables while quantifying causal effects over these partial structures to identify the root cause $R$.

\textbf{Formal Objective.} Given a normal dataset $D$ and an abnormal dataset $D^*$, our objective is to develop a fully data-driven approach able to: 
\begin{enumerate}
    \item Account for hidden variables during causal structure learning by estimating a PAG over $V \cup \{F\}$;
    \item Quantify the causal effects of the intervention $F$ on the candidate variables under the assumption of unobserved confounding; 
    \item Utilize these quantified causal effects to identify the true root cause $R$.
\end{enumerate}

\section{Methodology}
\label{sec:methodology}

\begin{figure}[htbp!]
    \centering
    \includegraphics[width=\columnwidth]{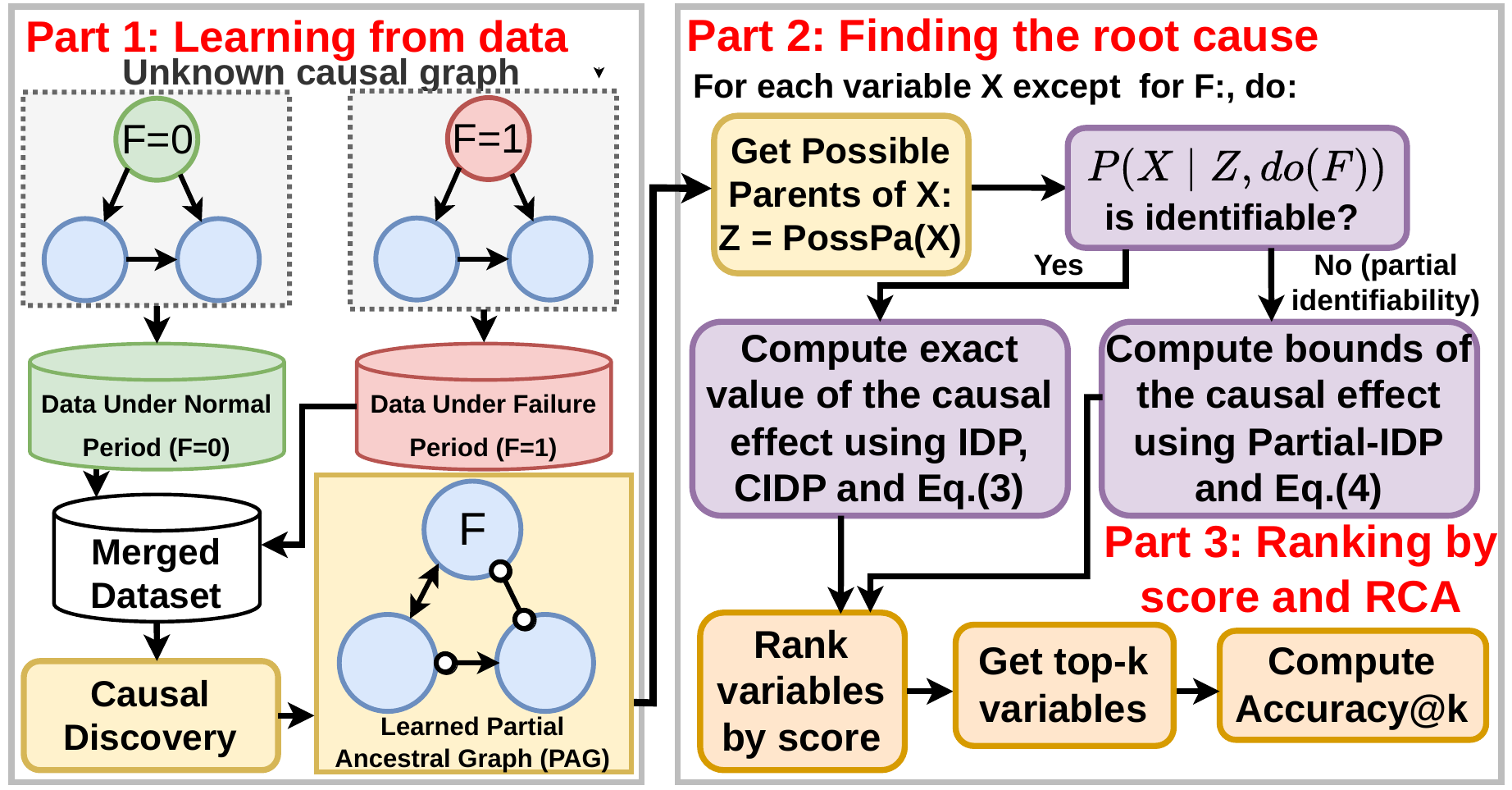}
        \caption{Flowchart of the proposed PAG-RCA algorithm. The true unknown causal graphs (gray) generate data under normal (green) and failure (red) periods. The merged dataset is used to learn the PAG with possible latent variables (yellow). Then, the proposed score algorithm is used to compute either the exact causal effects or their bounds (purple). Finally, the variables are ranked, and RCA accuracy is evaluated (orange).}
    \label{fig:flowchart}
\end{figure}

The flowchart of the proposed methodology is presented in Figure \ref{fig:flowchart}

\subsection{Assumptions}
To formalize our root cause ranking mechanism, we first establish the theoretical assumptions.

\textbf{Assumption 1 (Relaxation of Causal Sufficiency).} Unlike existing literature that assume all relevant system variables are observed (i.e. Causal Sufficiency), we allow for the presence of unmeasured latent variables, using PAGs to represent the causal uncertainties.

\textbf{Assumption 2 (Parametric Interventions).} Following established RCA formulations \cite{rcd_paper_2022, rcg_paper_2025, easy_rca_paper}, we model a system failure as a parametric (or soft) intervention. A parametric intervention alters the structural equation (the generative causal mechanism) of the root cause $R$ but does not sever its incoming edges. In other words, $Parents(R)$ remains unchanged, but the conditional distribution shifts.

\textbf{Assumption 3 (Extended Faithfulness).} We assume the joint distribution over the augmented variables $V \cup \{F\}$ is faithful to the underlying causal graph, implying that statistical independencies measured in the data perfectly map to d-separations in the graph, even in the presence of unobserved confounding.

\subsection{Scoring and Ranking Root Causes}
\label{subsec:scoring}

One of the main challenges in RCA is defining a scoring function that ranks system variables according to their likelihood of being the true source of a failure. The proposed scoring mechanism builds upon the current RCA literature with two primary goals: (1) utilizing rigorous causal effects to define and identify the root cause, and (2) extending this evaluation to account for unobserved latent variables by operating over PAGs. 

Frameworks such as EasyRCA \cite{easy_rca_paper} utilize the $do$-operator to evaluate mechanism shifts by computing the difference in the direct effect of a candidate node $X$ on a downstream node $Y$. They compare expectations between normal ($N$) and anomalous ($\bar{N}$) datasets, adjusting for a mediator set $\mathcal{W}$:

\begin{equation}
\begin{split}
Score_{EasyRCA}(X \rightarrow Y) = \mathbb{E}_{\bar{N}} [Y \mid do(X=x), \mathcal{W}=w] \\ - \mathbb{E}_{N} [Y \mid do(X=x), \mathcal{W}=w]
\end{split}
\end{equation}

By introducing the exogenous $F$-node to represent the operational state within a unified graph, the comparison between distinct $N$ and $\bar{N}$ datasets transforms into evaluating the system under the interventions $do(F=0)$ and $do(F=1)$. Consequently, the objective shifts from measuring the effect of $X$ on $Y$ to measuring the direct causal effect of the failure intervention $F$ on the candidate node $X$.

Methods such as RCD \cite{rcd_paper_2022} and RCG \cite{rcg_paper_2025} embed the $F$-node directly into the graphical structure. To handle structural uncertainties in MECs, RCG isolates the failure indicator $F$ from candidate nodes by evaluating the Conditional Mutual Information (CMI) given their possible parents:

\begin{equation}
Score_{RCG}(X) = CMI(F; X \mid PossPa_{\mathcal{G}}(X))
\end{equation}

While computationally efficient, this metric remains correlational and operates under the strict assumption of causal sufficiency. 

\textbf{Score under Full Identifiability.} The proposed framework extends upon these approaches by specializing interventional theory for partial graph-based systems. We formalize the root cause score as the expected divergence between the interventional mechanisms of the normal ($F=0$) and anomalous ($F=1$) states. We adopt the adjustment set logic from RCG to account for structural equivalence, defining the conditioning set as $Z = PossPa_{\mathcal{G}}(X) \setminus \{F\}$. However, we replace the associational CMI evaluation with an exact interventional calculation using the $do$-operator.

When the causal effect of $F$ on $X$ given $Z$ is point-identifiable over the learned PAG, we utilize the IDP and CIDP algorithms \cite{jaber2022causal} to compute the interventional distributions $P(X \mid do(F=f), z)$ solely from observational data. The point-identifiable root cause score is defined as:

\begin{equation}
\label{eq:rca_pag_score_id}
\begin{split}
Score(X) = \mathbb{E}_{z \sim Z} \Big[ \big| &P(X \mid do(F=0), z) - \\ 
&P(X \mid do(F=1), z) \big| \Big]
\end{split}
\end{equation}

In Equation \ref{eq:rca_pag_score_id}, the term $P(X \mid do(F=0), z)$ models the causal mechanism during normal operation, while $P(X \mid do(F=1), z)$ models the anomaly. The parametric intervention requires the generative mechanism of the true root cause to change, producing a divergence between these states that is directly captured by the score.

\textbf{Score under Partial Identifiability.} Structural uncertainties (e.g., undetermined edge marks $\circ$ in the PAG) or latent confounders often render the exact interventional distribution unidentifiable via standard CIDP. Instead of discarding these candidate nodes or relying on associational fallbacks, we account for partial identifiability. 

We employ the Partial CIDP algorithm \cite{bellot2024towards}, to derive lower and upper bounds for the unidentifiable effect directly from observational distributions. Let $\mathcal{I}_{f}^{(z)} = [L_f^{(z)}, U_f^{(z)}]$ represent the bounding interval for $P(X \mid do(F=f), z)$. 

To quantify the causal mechanism shift between these bounded spaces, we must measure the distance between the normal and anomalous intervals. We employ the Hausdorff distance \cite{maiseli2021hausdorff}, defining the generalized bounded score function as:

\begin{equation}
\label{eq:rca_pag_score_bound}
Score(X) = \mathbb{E}_{z \sim Z} \left[ \max \left( \left| L_0^{(z)} - L_1^{(z)} \right|, \left| U_0^{(z)} - U_1^{(z)} \right| \right) \right]
\end{equation}

By utilizing the Hausdorff distance within the expectation, the framework guarantees a non-zero score if the anomaly forces a contraction, expansion, or translation of the mechanism's probability space. With both Equations \ref{eq:rca_pag_score_id} and \ref{eq:rca_pag_score_bound} the proposed the proposed methodology can be used to evaluate RCA in both fully identifiable and partially identifiable scenarios.

\subsection{Proposed Algorithm}

This section details PAG-RCA (Algorithm \ref{alg:pag_rca}). It operates in two phases: causal structure discovery from the augmented datasets, followed by causal effect quantification and scoring. 

Given the continuous metrics intersecting the normal dataset $D$ and anomalous dataset $D^*$, we construct a joint dataset $D_{joint}$ and append the binary $F$-node. Utilizing the background knowledge $\mathcal{B}$ that $F$ acts as an exogenous intervention and therefore cannot have incoming edges \cite{rcg_paper_2025,rcd_paper_2022}, we apply the Fast Causal Inference (FCI) algorithm\footnote{using the Chi-squared conditional independence test with $\alpha=0.05$ for significance level} over $D_{joint}$ to learn the underlying PAG $\mathcal{P}$. 

Once $\mathcal{P}$ is learned, we evaluate every candidate metric $X \in V$. We extract its possible parents $Z$ from $\mathcal{P}$, excluding $F$ (i.e., $Z \gets PossPa_{\mathcal{P}}(X) \setminus \{F\}$). We then query the CIDP algorithm \cite{jaber2022causal} to derive the do-free expression $Exp$ for the causal effect of $F$ on $X$ given $Z$. 

If CIDP returns a valid expression, the causal effect is identifiable. We evaluate this expression over the observational data to obtain the exact interventional probabilities $P_0 = P(X \mid do(F=0), z)$ and $P_1 = P(X \mid do(F=1), z)$, and compute the root cause score directly via Equation \ref{eq:rca_pag_score_id}. Conversely, if CIDP returns a FAIL, the exact causal effect is not fully identifiable.\footnote{Because the CIDP algorithm is sound and complete, returning a FAIL mathematically guarantees that the conditional effect cannot be point-identified from the given PAG.} In this scenario, we must evaluate its causal bounds. The algorithm falls back to Partial-CIDP \cite{bellot2024towards} to derive the lower and upper intervals ($[L_0, U_0]$ and $[L_1, U_1]$) from observational data, scoring the candidate node using the bounded Hausdorff expectation (Equation \ref{eq:rca_pag_score_bound}). Finally, all candidate nodes are ranked in descending order, where the highest expected divergence indicates the most probable root cause.
\begin{algorithm}[!htbp]
\caption{PAG-RCA}
\label{alg:pag_rca}
\begin{algorithmic}[1]
\Statex \textbf{Input:} Normal data $D$, Anomalous data $D^*$
\Statex \textbf{Output:} Ranked list of root cause variables
\State $V \gets$ intersecting features of $D$ and $D^*$
\State $D_{joint} \gets D \cup D^*$ with appended $F$-node
\State $\mathcal{B} \gets$ restrict $F$ from having incoming edges
\State $\mathcal{P} \gets \text{FCI}(D_{joint}, \mathcal{B})$
\State $Scores \gets \emptyset$
\For{each $X \in V$}
    \State $Z \gets PossPa_{\mathcal{P}}(X) \setminus \{F\}$
    \State $Exp \gets \text{CIDP}(\mathcal{P}, F, X, Z)$ 
    \If{$Exp \neq \text{FAIL}$} \Comment{Effect is point-identifiable}
        \State $P_0 \gets \text{Eval}(Exp \mid F=0, D_{joint})$ 
        \State $P_1 \gets \text{Eval}(Exp \mid F=1, D_{joint})$ 
        \State $Scores[X] \gets \text{Eq. (\ref{eq:rca_pag_score_id}) using } P_0, P_1$
    \Else \Comment{Effect is unidentifiable; derive bounds}
        \State $[L_0, U_0] \gets \text{Partial-CIDP}(\mathcal{P}, F=0, X, Z)$
        \State $[L_1, U_1] \gets \text{Partial-CIDP}(\mathcal{P}, F=1, X, Z)$
        \State $Scores[X] \gets \text{Eq. (\ref{eq:rca_pag_score_bound}) using bounds}$
    \EndIf
\EndFor
\State \Return $Scores$ sorted in descending order
\end{algorithmic}
\end{algorithm}

\textbf{Illustrative example.} Consider the PAG $\mathcal{P}$ in Figure \ref{fig:running_example}c. To evaluate the causal effect of a failure on the candidate system metrics $V = \{X_1, X_2, X_3, X_4\}$, we first extract the set of possible parents $Z$ for each node, excluding the intervention node $F$. In a PAG, a node is a possible parent if the incident edge has no arrowhead pointing toward it. For $X_1$, the only valid incoming edge is from $F$; thus, excluding $F$ yields $Z_1 = \emptyset$, meaning we evaluate the marginal interventional distribution $P(X_1 \mid do(F))$. For $X_2$, the edges $X_1 \circ\rightarrow X_2$ and $X_2 \circ-\circ X_4$ designate $X_1$ and $X_4$ as possible parents, while the bidirected edge $X_2 \leftrightarrow X_3$ rules out $X_3$ as an ancestor. Therefore, $Z_2 = \{X_1, X_4\}$, and we must evaluate the conditional interventional distribution $P(X_2 \mid X_1, X_4, do(F))$. Similarly, the edge $X_1 \circ\rightarrow X_3$ makes $X_1$ the sole possible parent for $X_3$, yielding $Z_3 = \{X_1\}$ to evaluate $P(X_3 \mid X_1, do(F))$. Finally, the undirected uncertainty of $X_2 \circ-\circ X_4$ makes $X_2$ the sole possible parent for $X_4$, yielding $Z_4 = \{X_2\}$ to evaluate $P(X_4 \mid X_2, do(F))$. To compute these effects, the target node $X$, its corresponding parent set $Z$, and the intervention $do(F)$ are passed to the CIDP algorithm. CIDP evaluates the PAG to determine if the post-intervention probability $P(X \mid Z, do(F))$ can be reduced into a do-free expression computable from the joint observational dataset. If CIDP returns a valid expression, we compute the expected absolute difference between the normal state $P(X \mid Z, do(F=0))$ and the anomalous state $P(X \mid Z, do(F=1))$.

\section{Experiments}
\label{sec:experiments}

\paragraph{Implementation and Evaluation.} Following previous literature \cite{easy_rca_paper,rcd_paper_2022,rcg_paper_2025}, our data generation methodology centers on creating random causal graphs. We fix the number of nodes as 100. We then generated datasets for both normal and anomalous operational states, simulating the system failure by perturbing the data generation mechanism of a randomly selected root cause node. To evaluate our framework's robustness against hidden confounding, we extended the standard data generation process to include latent variables. During graph construction, we injected a predefined number of latent variables into the ground truth graph, assigning each to act as a confounder for multiple observed nodes. We generated datasets using this complete graph but removed the latent variables from the final datasets. This guarantees that the RCA algorithms only have access to the observed metrics, accurately simulating the unmeasured confounders prevalent in real-world environments. To ensure statistical robustness, each experimental configuration was repeated 100 times. For complete generation parameters and implementation details, please refer to the Extended version.

In our evaluations, we report \text{Accuracy@}$k$, a key metric used in the RCA literature for evaluating algorithm effectiveness. It is defined as the probability of successfully identifying the true root cause within the top $k$ ranked candidate nodes.

\paragraph{Baselines} We compare our proposed framework against the following baselines:

\begin{itemize}
    \item \textbf{RCD} \cite{rcd_paper_2022}: A fully data-driven approach employs a hierarchical and localized search strategy using conditional independence tests to detect the root cause.
    
    \item \textbf{RCG} \cite{rcg_paper_2025}: A data-driven method that extends the intervention framework by incorporating partial causal knowledge. RCG learns a CPDAG from observational data and applies marginal invariance tests to identify root causes.
    
    \item \textbf{BARO} \cite{baro_2024_paper}: A framework that constructs a causal graph via neural Granger causal discovery with contrastive learning and ranks the candidate nodes by executing a PageRank algorithm.
    
    \item \textbf{SMOOTH} \cite{orchard2026root}: A recent method that models anomalies as soft interventions and traverses the graph utilizing marginal anomaly scores. Unlike the purely data-driven methods described above, SMOOTH fundamentally assumes the true causal graph is known. Consequently, we include SMOOTH as a baseline to illustrate performance benchmarks when perfect causal knowledge is provided.
\end{itemize}

\subsection{Performance comparison}
\begin{figure}[htbp!]
    \centering
    \includegraphics[width=\columnwidth]{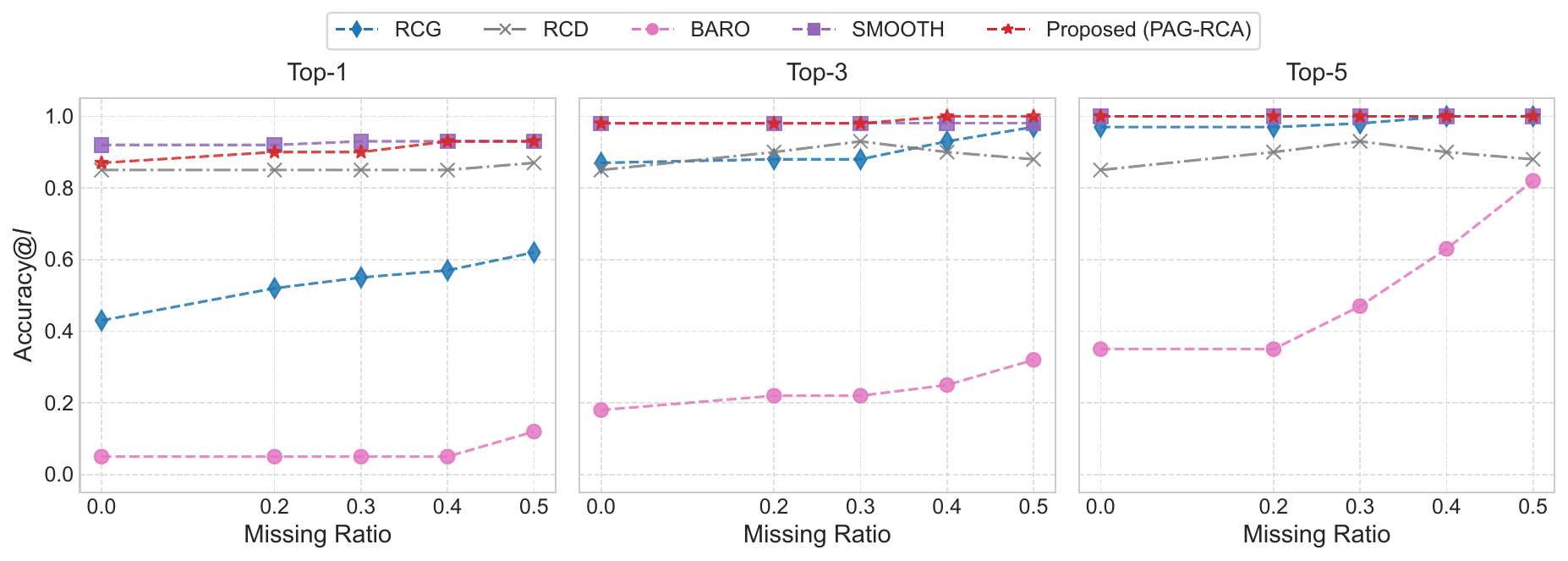}
        \caption{Top-$k$ accuracy evaluated under varying missing data ratios. }
    \label{fig:top_k_accuracy_missing_data}
\end{figure}

\paragraph{Missing Data.} Figure \ref{fig:top_k_accuracy_missing_data} presents the comparison, where missing ratio defines the fraction of randomly removed nodes from the observed dataset. Accuracy increases at higher ratios due to the reduced search space of candidate nodes. PAG-RCA achieves performance comparable to SMOOTH, a baseline that requires prior knowledge of the true causal graph. Among the data-driven approaches, PAG-RCA yields the highest accuracy. In Top-1 accuracy, PAG-RCA exceeds RCD and RCG, indicating that the interventional methodology isolates root causes under incomplete observability more effectively than prior frameworks using the F-node formulation.

\begin{figure}[htbp!]
    \centering
    \includegraphics[width=\columnwidth]{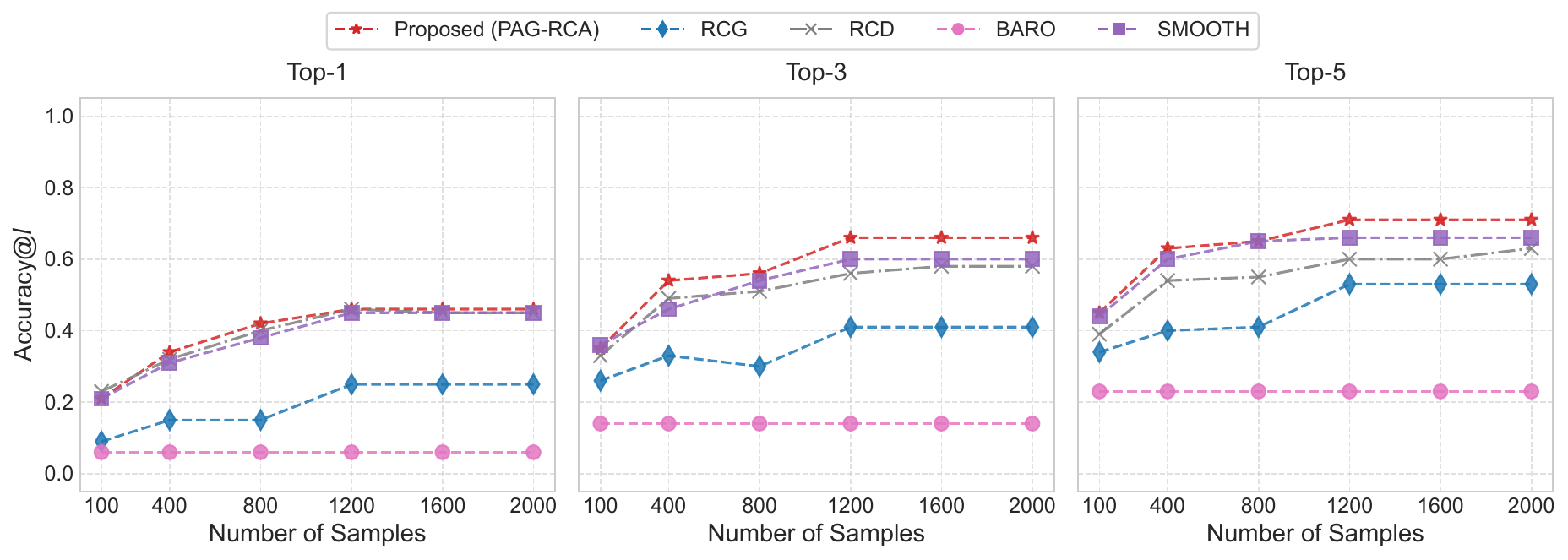}
        \caption{Top-$k$ accuracy evaluated across varying numbers of anomalous samples. }
    \label{fig:top_k_accuracy_num_samples}
\end{figure}

\paragraph{Number of Abnormal Samples.} Figure \ref{fig:top_k_accuracy_num_samples} evaluates performance across varying amounts of anomaly data. Accuracy for all data-driven techniques scales with sample size. PAG-RCA records the highest Top-1, Top-3, and Top-5 accuracies across the evaluated regimes. Baselines relying on associational metrics exhibit steeper performance degradation under limited anomaly data. PAG-RCA mitigates this by incorporating partial identifiability; when sample sparsity prevents point-identifiability of causal effects, the method computes analytical bounds from the $F$-node to candidate nodes to establish the final ranking.

\begin{table*}[t]
\centering
\resizebox{\textwidth}{!}{%
\begin{tabular}{l ccc ccc ccc ccc ccc}
\toprule
\multirow{2}{*}{\textbf{Method}} & \multicolumn{3}{c}{\textbf{Carts}} & \multicolumn{3}{c}{\textbf{Catalogue}} & \multicolumn{3}{c}{\textbf{Order}} & \multicolumn{3}{c}{\textbf{Payment}} & \multicolumn{3}{c}{\textbf{User}} \\
\cmidrule(lr){2-4} \cmidrule(lr){5-7} \cmidrule(lr){8-10} \cmidrule(lr){11-13} \cmidrule(lr){14-16}
& \textbf{Acc@1} & \textbf{Acc@3} & \textbf{Acc@5} & \textbf{Acc@1} & \textbf{Acc@3} & \textbf{Acc@5} & \textbf{Acc@1} & \textbf{Acc@3} & \textbf{Acc@5} & \textbf{Acc@1} & \textbf{Acc@3} & \textbf{Acc@5} & \textbf{Acc@1} & \textbf{Acc@3} & \textbf{Acc@5} \\
\midrule
BARO                    & 0.00 & 0.00 & \textbf{1.00} & 0.00 & 0.00 & \textbf{1.00} & 0.00 & \textbf{1.00} & \textbf{1.00} & \textbf{0.80} & \textbf{0.80} & \textbf{1.00} & 0.00 & \textbf{1.00} & \textbf{1.00} \\
RCD                     & 0.20 & 0.60 & 0.60 & 0.20 & 0.20 & 0.40 & 0.20 & 0.60 & 0.60 & 0.60 & \textbf{0.80} & 0.80 & \textbf{0.40} & 0.40 & 0.40 \\
RCG                     & 0.20 & 0.60 & \textbf{1.00} & 0.40 & 0.80 & \textbf{1.00} & 0.20 & 0.60 & 0.80 & 0.20 & \textbf{0.80} & \textbf{1.00} & 0.00 & 0.80 & \textbf{1.00} \\
\midrule
\textbf{PAG-RCA (Ours)} & \textbf{1.00} & \textbf{1.00} & \textbf{1.00} & \textbf{0.60} & \textbf{1.00} & \textbf{1.00} & \textbf{0.60} & 0.80 & \textbf{1.00} & 0.20 & 0.20 & 0.40 & 0.00 & \textbf{1.00} & \textbf{1.00} \\
\bottomrule
\end{tabular}%
}
\caption{RCA performance on the Sock-shop dataset for memory leak failures. Best results are in bold.}
\label{tab:sock_shop_combined}
\end{table*}

\begin{table}[t]
\centering
\resizebox{\columnwidth}{!}{
\begin{tabular}{l ccc ccc}
\toprule
\multirow{2}{*}{\textbf{Method}} & \multicolumn{3}{c}{\textbf{IEEE 14-Bus}} & \multicolumn{3}{c}{\textbf{IEEE 30-Bus}} \\
\cmidrule(lr){2-4} \cmidrule(lr){5-7}
& \textbf{Acc@1} & \textbf{Acc@3} & \textbf{Acc@5} & \textbf{Acc@1} & \textbf{Acc@3} & \textbf{Acc@5} \\
\midrule
SMOOTH                 & 0.00 & 0.28 & 0.66 & 0.00 & 0.29 & 0.36 \\
BARO                   & 0.28 & 0.82 & \textbf{1.00} & 0.21 & 0.43 & 0.43 \\
RCD                    & 0.46 & 0.62 & 0.62 & 0.14 & 0.29 & 0.29 \\
RCG                    & 0.30 & 0.58 & 0.62 & 0.21 & 0.29 & 0.43 \\
\midrule
\textbf{PAG-RCA (Ours)} & \textbf{0.48} & \textbf{0.84} & 0.90 & \textbf{0.43} & \textbf{0.79} & \textbf{0.86} \\
\bottomrule
\end{tabular}
}
\caption{RCA performance on power grid cascading failure events under the IEEE14 and IEEE30 test cases. Best results are in bold.}
\label{tab:cascading_failures_combined}
\end{table}

\begin{figure}[htbp!]
    \centering
    \includegraphics[width=\columnwidth]{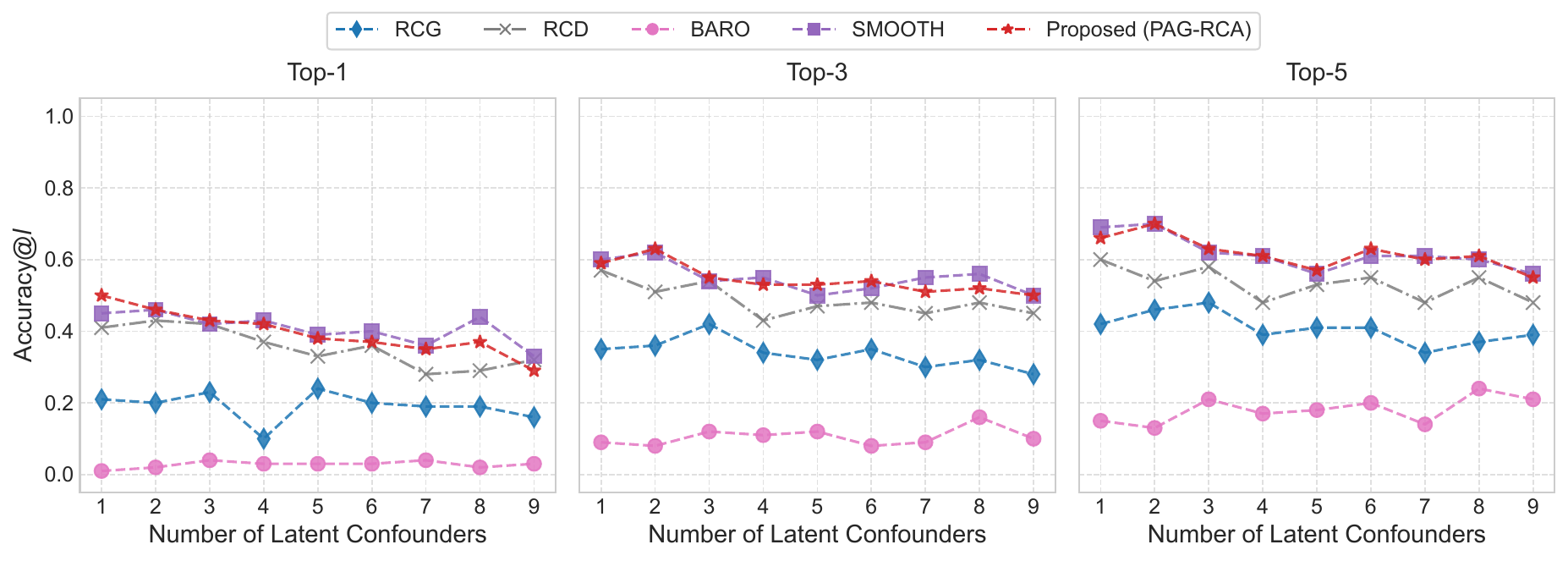}
        \caption{Top-$k$ accuracy evaluated across varying numbers of latent confoundes (hidden common causes).}
    \label{fig:top_k_accuracy_latent_confounders}
\end{figure}

\paragraph{Latent Variables.} Figure \ref{fig:top_k_accuracy_latent_confounders} presents the results under unobserved confounding. Introducing latent variables decreases accuracy across all methods compared to fully observable environments. SMOOTH relies on prior knowledge of the true graph, including the exact placement of latent confounders. Without this prior knowledge, PAG-RCA achieves accuracy levels comparable to SMOOTH, while the performance of the other data-driven models declines. Isolating the exact Top-1 root cause is mathematically constrained under dense latent confounding. In this setting, PAG-RCA yields higher Top-3 and Top-5 accuracies than the data-driven baselines, isolating a bounded subset of candidate components to guide repair strategies.

\section{Real Data}
\label{sec:case_study}

\paragraph{Sock-shop Dataset.} To evaluate the proposed framework in a real-world environment, we utilize the Sock-shop microservice dataset, widely adopted as a standard benchmark in the RCA literature \cite{orchard2026root,cyclic_rca_paper_2026,rcg_paper_2025,rcd_paper_2022}. It is an e-commerce replica comprising 13 interacting microservices deployed in separate containers, generating multivariate telemetry such as CPU utilization, memory usage, and HTTP latency.

\paragraph{Cascading Failures in Power Grids.} Power grids are highly interdependent critical infrastructures where a single localized fault (e.g., a transmission line outage) can trigger blackouts known as cascading failures. Due to their complex graph structures, power grids and their failure dynamics have recently emerged as benchmarks for evaluating machine learning models \cite{varbella2024powergraph}. In this domain, the RCA objective translates to identifying the initial line outage that triggered the cascade. To evaluate our framework, we simulated cascading failures over the standard IEEE 14-bus and 30-bus test systems, which comprise 20 and 41 physical lines, respectively. The data generation mechanism utilizes the enhanced OPA model \cite{park2022enhanced}, simulating stochastic daily operations and faults over a 2,000-day period. The dataset is provided in our previous work \cite{caetano2026spatiotemporal} and in the Extended version.

Tables \ref{tab:sock_shop_combined} and \ref{tab:cascading_failures_combined} present results for each dataset and compare them with previously detailed baselines. In the Sock-shop dataset, the framework achieves localization accuracy (Acc@1) of 1.00 in the \textit{Carts} service and 0.60 in \textit{Catalog} and \textit{Order}, reducing spurious correlations that affect methods like RCD and RCG. Unidentifiability of causal effects in the \textit{Payment} and \textit{User} services lowers Acc@1. The bounded Hausdorff expectation isolates the root cause within the Top-3 candidates, showing the utility of partial identifiability under latent confounding. This evaluation also applies to cascading failures, where PAG-RCA records the highest Acc@1 for the IEEE 14-bus (0.48) and IEEE 30-bus (0.43) systems. 

PAG-RCA outperforms data-driven baselines in both microservice and cascading-failure contexts by relaxing the assumption of causal sufficiency. It outperforms F-node frameworks such as RCD and RCG, which rely on associational metrics such as CMI, as well as BARO, which uses time-series correlation and change-point detection. This highlights the advantages of using causality in RCA. Additionally, PAG-RCA excels over SMOOTH, a baseline that assumes a fully known causal graph. This showcases the effectiveness of incorporating techniques such as partial identifiability and latent variables into causal-based RCA.

\section{Conclusion}

RCA is critical for maintaining system reliability, yet existing frameworks assume causal sufficiency, a fundamental limitation that fails to account for the unobserved latent confounders prevalent in real-world environments. This paper introduces PAG-RCA, a data-driven RCA framework that models system failures as parametric interventions over PAGs, addressing the problem of latent variables for the first time in RCA literature. Our methodology overcomes the limitations of traditional associational metrics and strict assumptions of causal sufficiency. When exact causal effects are unidentifiable in the PAG, PAG-RCA utilizes partial identification and causal bounding to ensure RCA under these conditions. Evaluations on synthetic datasets, microservice benchmarks, and power grid cascading failures show that our approach outperforms state-of-the-art baselines, especially in scenarios with data scarcity and latent confounding. 

\section*{Acknowledgments}

This work was partially fomented by São Paulo Research Foundation (FAPESP), grants 2021/12220-1, 2023/07634-7 and 2024/08485-8.

\bibliography{aaai2027}


\end{document}